\documentclass[10pt,letterpaper]{article}
\usepackage[top=0.85in,left=1.55in,footskip=0.75in,marginparwidth=2in]{geometry}

\usepackage{authblk}
\usepackage{times}
\usepackage{hyperref}
\usepackage{url}
\usepackage{color}
\usepackage{hyperref}
\usepackage{amsmath}
\usepackage{amsfonts}
\usepackage{amssymb}
\usepackage{dsfont}
\usepackage{graphicx}
\usepackage{wrapfig}
\usepackage{amsthm}
\usepackage{natbib}
\usepackage{color}
\usepackage{mathtools}
\usepackage[]{algorithm2e}

\newcommand{\sgn}{\operatorname{sgn}}

\title{Defending against Adversarial Images using Basis Functions Transformations}
\author[1]{Uri Shaham\footnote{Equal contribution.}}
\newcommand\CoAuthorMark{\footnotemark[\arabic{footnote}]} 
\author[1]{James Garritano\protect\CoAuthorMark}
\author[1]{Yutaro Yamada}
\author[1]{Ethan Weinberger}
\author[2]{Alex Cloninger}
\author[3]{Xiuyuan Cheng}
\author[1]{Kelly Stanton}
\author[1]{Yuval Kluger}
\affil[1]{Yale University}
\affil[2]{University of California, San Diego}
\affil[3]{Duke University}

\begin{document}

\maketitle

\begin{abstract}
We study the effectiveness of various approaches that defend against adversarial attacks on deep networks via manipulations based on basis function representations of images. Specifically, we experiment with low-pass filtering, PCA, JPEG compression, low resolution wavelet approximation, and soft-thresholding. We evaluate these defense techniques using three types of popular attacks in black, gray and white-box settings. 
Our results show JPEG compression tends to outperform the other tested defenses in most of the settings considered, in addition to soft-thresholding, which performs well in specific cases, and yields a more mild decrease in accuracy on benign examples. 
In addition, we also mathematically derive a novel white-box attack in which the adversarial perturbation is composed only of terms corresponding a to pre-determined subset of the basis functions, of which a ``low frequency attack'' is a special case.
\end{abstract}

\section{Introduction}
\label{sec:introduction}

In the past five years, the areas of adversarial attacks~\citep{szegedy2013intriguing} on deep learning models, as well as defenses against such attacks, have received significant attention in the deep learning research community~\citep{yuan2017adversarial, akhtar2018threat}.

Defenses against adversarial attacks can be categorized into two main types. 
Approaches of the first type modify the net training procedures or architectures, usually in order to make the net compute a smooth function; see, for example~\citep{shaham2015understanding, gu2014towards, cisse2017parseval, papernot2016distillation}.
Defenses of the second type leave the training procedure and architecture unchanged, but rather modify the data, aiming to detect or remove adversarial perturbations often by smoothing the input data.
For example,~\cite{guo2017countering} applied image transformations, such as total variance minimization and quilting to smooth input images.
~\citet{dziugaite2016study, das2017keeping} proposed to apply JPEG compression to input images before feeding them through the network. 
Closely related approaches were taken by~\cite{akhtar2017defense}, by applying the Discrete Cosine Transform (DCT) and by~\citet{bhagoji2017dimensionality, hendrycks2017early, li2016adversarial}, who proposed defense methods based on principal component analysis (PCA).
De-noising using PCA, DCT and JPEG compression essentially works by representing the data using a subset of its basis functions, corresponding to the first principal components, in case of PCA, or low frequency terms, in case of DCT and JPEG. A similar idea can be applied by low-pass Fourier filtering and wavelet approximation.

In this manuscript, we continue in this direction, by investigating various defenses based on manipulations in a basis function space. Specifically, we experiment with low-pass filtering, wavelet approximation, PCA, JPEG compression, and soft-thresholding of wavelet coefficients. 
We apply each of these defenses as a pre-processing step on both adversarial and benign images, on the Inception-v3 and Inception-v4 networks.
The defenses are applied only at test time (so that we do not re-train or change the publicly available network weights), and for each defense we evaluate its success at classifying adversarial images, as well as benign images.
We evaluate these defenses in black, gray and white-box settings, using three types of popular attacks.
In a black-box setting the attacker has no access to the gradients and no knowledge of the pre-processing procedure.
In a gray-box setting, the attacker has access to the gradients of the attacked network, however he does not have any knowledge of defenses being applied.
In a white-box setting, the attacker has access to the gradients, as well as full knowledge of the pre-processing procedure taking place.
Our results show that JPEG compression performs consistently as well as and often better than the other defense approaches in defending against adversarial attacks we experimented with, across all types of adversarial attacks in black-box and gray-box settings, while also achieving high performance under two different white-box attack schemes.
Soft-thresholding has the second best performance in defending against adversarial attacks, while outperforming JPEG compression on benign images.

In addition, we also mathematically derive a novel type of attack, in which the adversarial perturbation affects only terms corresponding to pre-specified subset of the basis functions. We call this approach a ``filtered gradient attack''. Several cases of special interest of this attack are when this subset contains only low frequency basis functions, coarse level wavelet functions, or first principal components.

The remainder of this manuscript is organized as follows. 
In Section~\ref{sec:attacks} we review the attacks used in our experiments, whereas in Section~\ref{sec:defenses} we review our defense approaches.
Our experimental results are provided in Section~\ref{sec:experiments}. 
Section~\ref{sec:conclusions} briefly concludes the manuscript.


\section{Attacks}
\label{sec:attacks}

\begin{figure*}[t]
  \centering
\includegraphics[width=13.8cm]{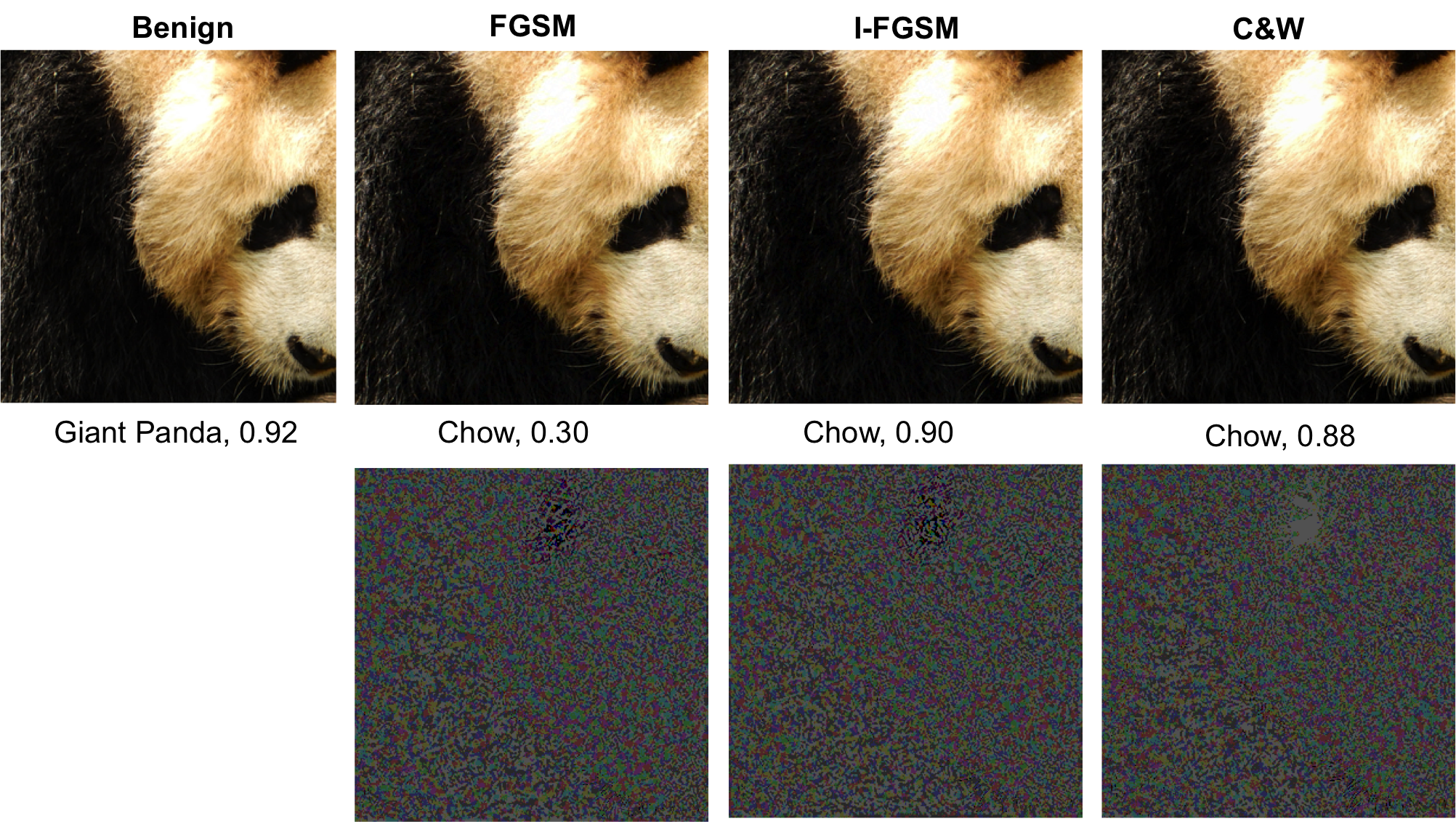}
\caption{Top: a benign example, FGSM, I-FGSM and C\&W adversarial examples, along with their top 1 prediction and confidence score. Bottom: the corresponding adversarial perturbations, multiplied by 50 for visibility.}
\label{fig:diffs}.
\end{figure*}

In this manuscript we experiment with three popular adversarial attacks.
\newline
\textbf{Fast Gradient Sign Method (FGSM) }~\citep{goodfellow2014explaining} is a fast method to generate an adversarial perturbation, where the perturbation is performed by either adding or subtracting a fixed amount from each image pixel, depending on the sign of the corresponding entry in the gradient of the loss function with respect to the image. Specifically, for an image $x$ with true label $y$, the adversarial image is $x'=  \text{clip}(x+\Delta_x)$, and the adversarial perturbation $\Delta_x$ is given by  
$$\Delta_x = \epsilon \cdot\sgn \left(\nabla_xJ_{\theta,y}(x)\right),$$
where $\sgn$ performs elementwise and $J_{\theta,y}(x)$ denotes the loss of a network with parameter vector $\theta$ on $(x,y)$, viewed as a function of $x$ (i.e., holding $\theta$ and $y$ fixed).
This perturbation can be derived from a first-order Taylor approximation of $J$ near $x$,
$$\max_{\Delta_x}J_{\theta,y}(x + \Delta_x) \approx J_{\theta,y}(x) + \langle \nabla_xJ_{\theta,y}(x),\Delta_x\rangle, $$
maximized by choosing $\Delta_x$ from a $\ell_\infty$ ball of radius $\epsilon$~\citep{shaham2015understanding}. The radius $\epsilon$ corresponds to the magnitude of the perturbation. 
\newline
\textbf{Iterative Fast Gradient Sign Method (I-FGSM) }~\citep{kurakin2016adversarial} works by repeated applications of the FGSM perturbation
$$x^{(m)} = \text{clip}\left(x^{(m-1)} +  \epsilon \sgn\cdot \nabla_xJ_{\theta,y}(x^{(m-1)})\right),$$
and setting the adversarial image to be $x'=x^{(M)}$, the output of the last iteration.
\newline
\textbf{Carlini-Wagner (C\&W) } is a family of attack methods, which typically yield perturbations of small magnitude.  
They utilize a margin which enables one to generate adversarial examples which are subsequently misclassified with high confidence.
Following~\cite{guo2017countering}, we use the  C\&W $\ell_2$ variant
\begin{align}
&\min_{x'} [\|x-x'\|^2 +\notag\\
&\lambda_f \max \left(-\kappa, Z(x')_{f(x)} - \max\left\{Z(x')_c:c\neq f(x)\right\} \right),\notag
\end{align}
where $\kappa$ is a margin, which will be explained later, $c$ is a class index, $f(x)$ is the network prediction on $x$, $Z(x)$ is the logit (i.e., pre-activation of the softmax output layer) for $x$ and $\lambda_f$ is a trade-off parameter. 
The left-most $\max$ term of the C\&W loss corresponds to the most probable class which is not the originally predicted one.
For $\kappa=0$, the left part of the loss is minimized when there is a class $c$ which is not the original predicted class $f(x)$, and whose logit is at least as big as the logit of the true class. 
Increasing the margin $\kappa$ requires that the gap between the logits increases correspondingly, resulting in a high confidence adversarial example. In our experiments we use $\kappa=0$.
The fact that C\&W perturbations are typically small in magnitude is a result of minimizing the squared difference $\|x-x'\|^2$. Unlike FGSM and I-FGSM examples, C\&W examples are much slower to generate, requiring applying an optimizer for each adversarial example.

Figure~\ref{fig:diffs} shows a benign image, adversarial images generated using the FGSM, I-FGSM and C\&W methods and the corresponding perturbations generated by those methods. 
The adversarial perturbations were generated using the Inception-v3 network.


\section{Defenses}
\label{sec:defenses}

\begin{figure*}[t]
  \centering
\includegraphics[width=11.1cm]{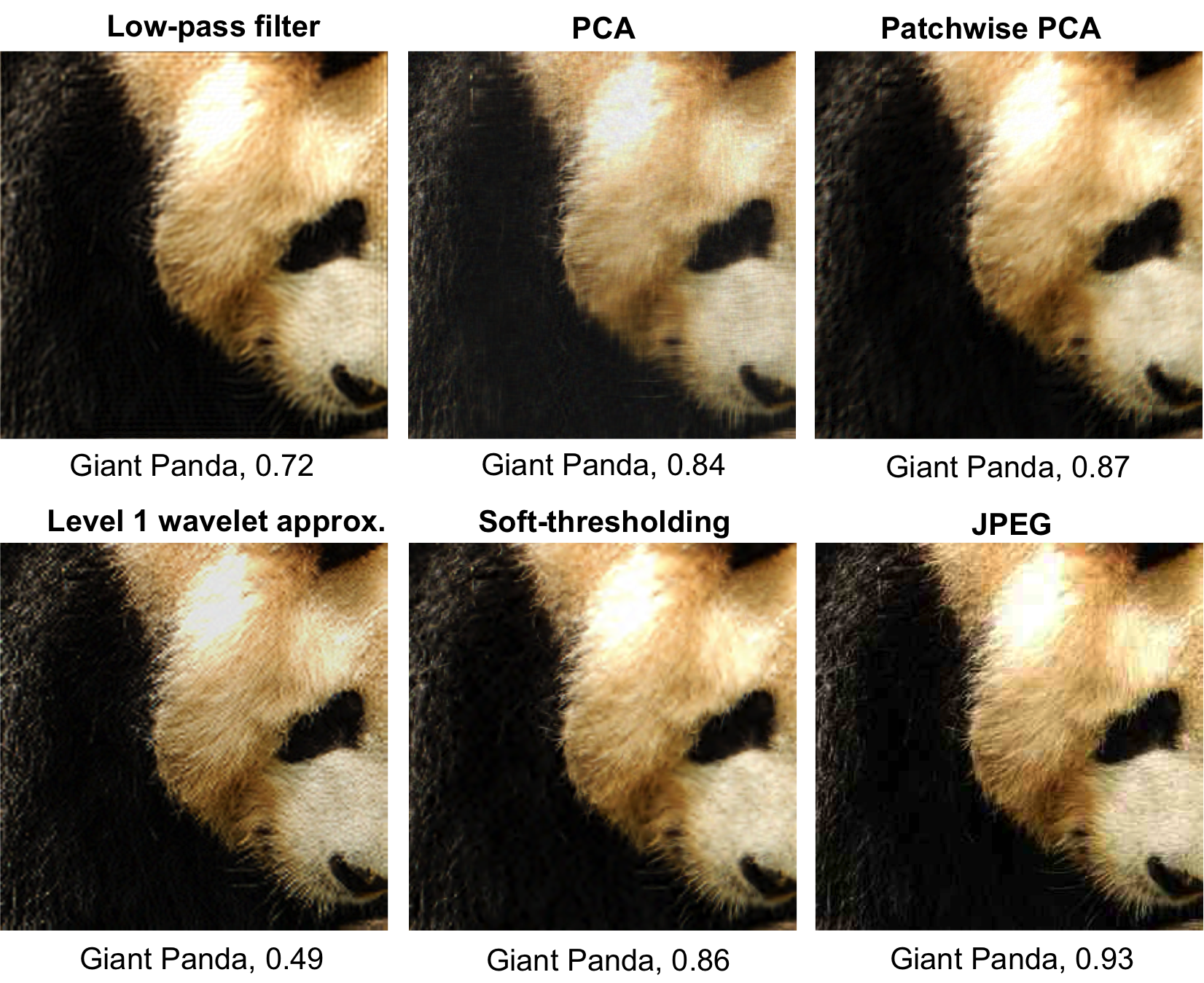}
\caption{Examples of the defense methods used in this manuscript, applied on the FGSM adversarial image in Figure~\ref{fig:diffs}.}
\label{fig:defenses_demo}.
\end{figure*}

We experiment with several defense methods, all of which operate by performing manipulations in basis function spaces. Below we describe each of the defenses used in our experiments. Further technical details are given in Section~\ref{sec:technicalDetails}.
\newline
\textbf{Low-pass filtering: }
The discrete Fourier transform of a two-dimensional signal (e.g., an image) $x_{n_1,n_2},\; n_i=1,\ldots N_i$ is defined by
\begin{equation}
X_{\bf k} = \sum_{{\bf n}=0}^{\bf N-1}x_n e^{-2\pi i \langle{\bf k}, ({\bf n/N})\rangle} ,\notag
\end{equation}
where ${\bf n} =(n_1,n_2)$, ${\bf k} =(k_1,k_2)$, ${\bf N-1} =(N_1-1,N_2-1)$ and ${\bf n/N} =(n_1/N_1,n_2/N_2)$.
Low pass-filtering is performed by obtaining the Fourier representation of the image, followed by element-wise multiplication of the Fourier coefficients $X_{\bf k}$ with a bump function, so that high frequencies are depressed, and lastly converting the signal back to pixel space using the inverse transform.
\newline
\textbf{PCA:}
PCA de-noising represents a given matrix by a low-rank approximation of it, while keeping as much variance as possible in the original data matrix (whose rows are viewed as data points and columns as input features).
This is done by obtaining the principal components of a matrix, representing the data in the PC space, discarding all but the $k$ leading principal directions and mapping the data back to its input space. Mathematically, this procedure is formalized by 
$$X_\text{pca} = XUU^T,$$
where $X$ is $n\times d$ matrix where each row corresponds to a data point, and $U$ is a $d\times k$ matrix containing the leading $k$ eigenvectors of the $d\times d$ covariance matrix $\frac{1}{n}(X-\bar{X})^T(X-\bar{X})$.
Rather than computing the PCA on the entire image dataset, as was done by~\citet{bhagoji2017dimensionality, hendrycks2017early, li2016adversarial}, we compute the principal components for each image separately, in two different ways:
\begin{itemize}
\item Viewing the image  as a matrix of size $n_\text{rows} \times n_\text{columns}$, i.e., where rows are considered as data points, and performing PCA denoising on that matrix.
\item We cut patches from each image, re-shape each patch to a vector and obtain a matrix whose rows are the patch vectors. We then perform PCA denoising on that matrix.
\end{itemize}
in both cases, we apply the denoising on each color channel separately.
\newline
\textbf{Wavelet approximation: }
Unlike complex exponentials, the Fourier basis functions, wavelet basis functions are localized. 
Wavelet basis on $\mathbb{R}^2$ is an orthonormal collection $\{\psi_{k,b}\}$ of zero-mean functions, created from a bump function $\varphi(x):\mathbb{R}^2\rightarrow \mathbb{R}$ (``father wavelet'') via 
$$\psi(x) = \varphi(x) - 2^{-1}\varphi\left(2^{-\frac{1}{2}}x\right) $$
and
$$\psi_{k,b}(x) = 2^{\frac{k}{2}}\psi\left(2^{\frac{k}{2}}(x-b)\right).$$
The index $k$ corresponds to the level of approximation (via the width of the bump) and $b$ to the shift.
Wavelet decomposition of a real-valued signal $f:\mathbb{R}^2 \rightarrow \mathbb{R}$ is represented as sequences of coefficients, where the $k$th sequence describes the change of the signal at the $k$th level, and $k=1,2,\ldots$. 
$f$ is then represented as
\begin{equation}
f(x) = \sum_{k\in \mathbb{Z}}\sum_{b\in \mathbb{Z}}\langle f, \psi_{k,b}\rangle \psi_{k,b}(x).\notag
\end{equation}
Discrete wavelet transform is a wavelet transform where the wavelets are discretely sampled. 
Since wavelet functions are localized, a wavelet basis is often better than the Fourier basis at representing non-smooth signals, yielding sparser representations. 
For 2D images, level $k$ wavelet approximation results in an approximation image of resolution which is coarser as $k$ grows, containing $2^{-2k}$ of the pixels of the original image. 
To resize the approximation image back to the original size, we use bi-cubic interpolation, implemented via Matlab's \textit{imresize} function.
\newline
\textbf{Soft-thresholding: }
~\cite{donoho1994ideal} consider a setting where one wishes to reconstruct a discrete signal $f$ from a noisy measurement $x$ of it, where
\begin{equation}
x_i = f(t_i) + \sigma z_i,\notag
\end{equation}
$z_i$s are iid standard Gaussian random variables, $i=1,\ldots,n$ and $t_i = i/n$.
They propose to de-noise $x$ using soft-thresholding of its wavelet coefficients, where the soft-thresholding operator is defined by
$$\eta(c) = \sgn(c) \max\{0, |c|-t\} $$
and t is a threshold, usually chosen to be
$$t=\sigma \sqrt{2\log n}.$$
A classical result by~\cite{donoho1995noising} proves that such de-noising is min-max optimal in terms of $\ell^2$ distance $\|\hat{f}-f \|$ between the de-noised signal $\hat{f}$ and the original one, while keeping $\hat{f}$ at least as smooth as $f$.
\newline
\textbf{JPEG compression: }
JPEG is a lossy compression that utilizes DCT and typically removes many high frequency components, to which human perception is less sensitive.
Specifically, JPEG compression consists of the following steps:
\begin{enumerate}
\item Conversion of the image from RGB format to $YC_bC_r$ format, where the $Y$ channel represents Luminance and $C_b,C_r$ channels represent chrominance. \item Down-sampling of the chrominance channels.
\item Splitting of the image to $8\times 8$ blocks and applying 2D DCT to each block. This is done for each channel separately.
\item Quantization of the frequency amplitudes, achieved by dividing each frequency term by a (different) constant and rounding it to the nearest integer. As a result, many high frequency components are typically set to zero, and others are shrinked. The amount of compression is governed by a user-specified quality parameter, defining the reduction in resolution.
\item Lossless compression of the block data.
\end{enumerate}
The lossy elements of JPEG compression are the down-sampling (step 2) and the quantization (step 3), where most of the compression is achieved.
JPEG defense was applied by~\citet{dziugaite2016study, das2017keeping,guo2017countering}.
\newline
In the case of color images, Fourier and wavelet transforms are typically applied on each color channel separately.
Figure~\ref{fig:defenses_demo} demonstrates the above defense methods on the panda image of Figure~\ref{fig:diffs}.


\section{Experiments}
\label{sec:experiments}

\subsection{Setup and Technical Details}
\label{sec:technicalDetails}
Our experiments were performed on the publicly available dataset from the NIPS 2017 security competition\footnote{\url{https://www.kaggle.com/c/nips-2017-defense-against-adversarial-attack/data}}, containing 1000 images, as well as a trained Inception-v3 model. 
All attacks were carried out using Cleverhans\footnote{\url{https://github.com/tensorflow/cleverhans/}}. For C\&W we used with $\kappa=0$ and $\lambda_f=0.02$, and similarly to~\citep{guo2017countering}, the perturbations were multiplied by constant $\epsilon \ge 1$ to alter their magnitude. 
FGSM and I-FGSM attacks were performed with $\epsilon \in [0.005,0.09]$.
The parameters of each of the defenses were selected to optimize the performance of the defense in a gray-box setting. 
Specifically, the low-pass filtering was applied by multiplying the Fourier coefficients of each color channel with a circle with radius of 65; PCA was performed by retaining the largest 36 principal components of each image; Patchwise PCA was performed on patches of size $13\times 13$, and retaining the largest 13 principal components;
JPEG compression was performed by setting the quality parameter to 23\%.
Wavelet approximation was performed in Matlab using the \textit{appcoef2} command; soft-thresholding was done using Matlab's \textit{ddencmp} and \textit{wdencmp}; for both approaches we used the bi-orthogonal \textit{bior3.1} wavelet.
In all experiments the adversarial examples were generated using the Inception-v3 network. We did not perform any re-training or fine-tuning of the net. 
All defenses were applied as test time pre-processing methods.
Our codes are available at~\url{https://github.com/KlugerLab/adversarial-examples}.

\subsection{Evaluation}
Following~\cite{guo2017countering}, for each defense we report the top-1 accuracy versus the normalized $\ell_2$ norm of the perturbation, defined by 
$$\frac{1}{n}\sum_{i=1}^n\frac{\|x_i-x_i' \|_2}{\|x_i \|_2}, $$
where $x$ and $x'$ denote benign and adversarial examples respectively, and $n$ is the number of examples. 

\subsection{Black-Box Setting}

\begin{figure}[t!]
  \centering
\includegraphics[width=6.3cm]{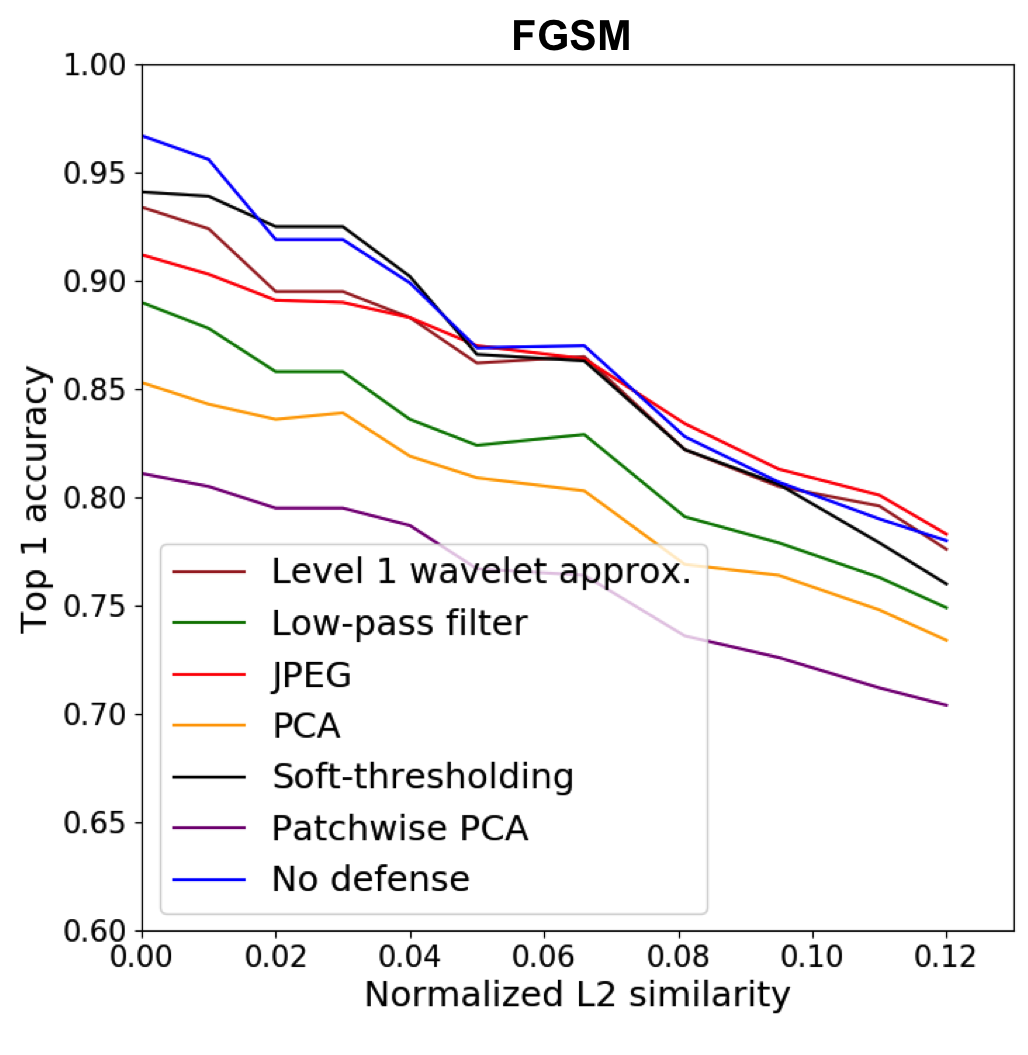}
\includegraphics[width=6.3cm]{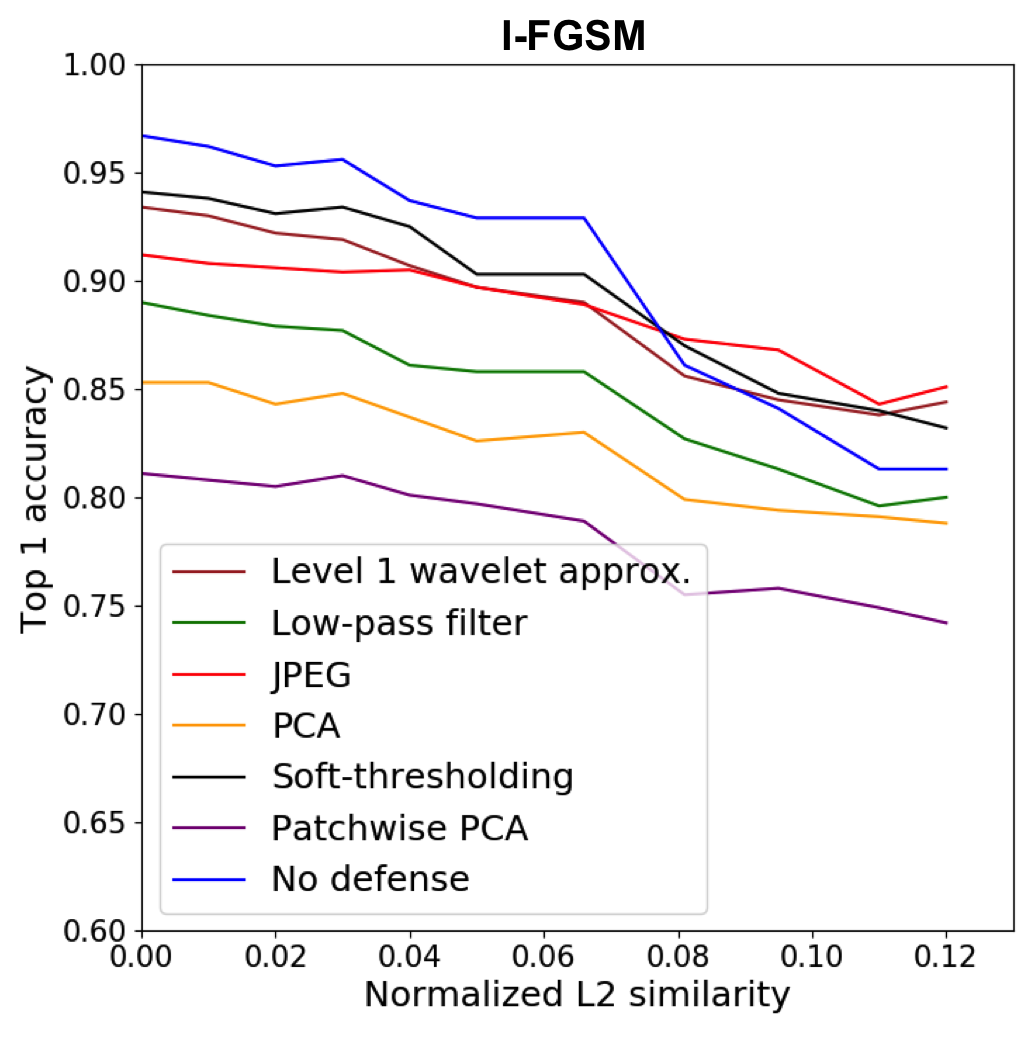}
\includegraphics[width=6.3cm]{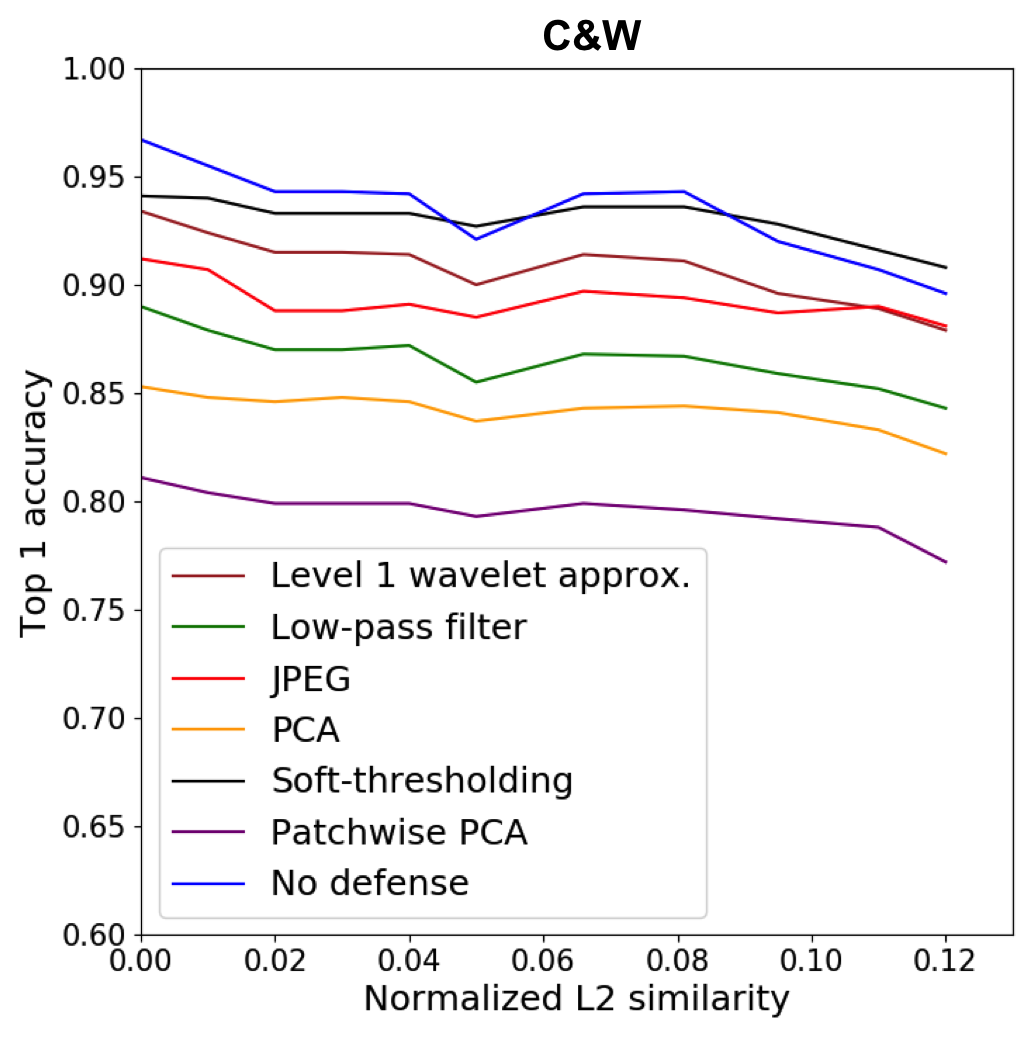}
\caption{Top 1 accuracy of each of the defense methods investigated in this manuscript on FGSM, I-FGSM and C\&W adversarial examples, in a black-box setting. The examples were generated using Inception-v3 and tested on Inception-v4.}
\label{fig:defenses_black}
\end{figure}

In this setting, the attacker has no access to the gradients of the target network. Rather, the attack is based on the transferability property of adversarial examples~\citep{liu2016delving, papernot2016transferability}.
Specifically, we generated adversarial examples generated for Inception-v3, applied each of the defenses (separately) on each adversarial example, fed them into Inception-v4, and measured the top 1 accuracy.
Figure~\ref{fig:defenses_black} shows the performance of each defense method as a function of the normalized $\ell_2$ norm of the perturbation, for each of the attack methods.
Overall, we found that in our experiment setup, transferability actually requires fairly large perturbations comparing to a gray-box setting (see Section~\ref{sec:gray})\footnote{We found that the largest normalized $\ell_2$ we consider in the black-box case corresponds to about $2/3$ of the one used by~\citet{liu2016delving}. We chose not to use larger perturbations as these become fairly noticeable to a human eye, and hence less adversarial.}.
Adversarial examples with perturbations of normalized $\ell_2$ norm below 0.08 generally only yield a modest decrease in accuracy. 
Consequently, all the tested defenses are ineffective against small perturbations in this setting.
JPEG denoising becomes effective around 0.08 against FGM and I-FGM attacks, where it outperforms all other defenses, however does not perform well against C\&W attacks. 
Soft-thresholding becomes effective against I-FGM and C\&W attacks for norms above 0.08. 
Low-pass filtering and the PCA denoising methods do not perform well against any attack.
In addition, Table~\ref{tab:defenses_black} displays the performance of each of the defenses on benign examples using Inception-v4.
\begin{table}
\centering
\scalebox{0.8}{
\begin{tabular}{ |l|l|}
  \hline
  Defense                                  & Top 1 accuracy    \\\hline\hline
  No defense                               &  0.967            \\\hline
  Low-pass filter                          &  0.89             \\\hline
  PCA                                      &  0.853            \\\hline
  JPEG                                     &  0.912            \\\hline
  Level 1 wavelet approximation            &  0.933            \\\hline
  Soft-thresholding                        &  0.941            \\\hline
  Patchwise PCA                            &  0.811            \\\hline
\end{tabular}
}
\caption{Performance of the defenses on benign examples on Inception-v4. The performance of each defense on adversarial examples is shown in Figure~\ref{fig:defenses_black}.}
\label{tab:defenses_black}
\end{table}
As can be seen, the wavelet-based methods (level 1 approximation and soft-thresholding yield the smallest decrease in accuracy on benign examples, followed by JPEG. Low-pass filtering and PCA methods yield a more significant decrease in accuracy.


\subsection{Gray-Box Setting}
\label{sec:gray}

\begin{figure}[t!]
  \centering
\includegraphics[width=6.3cm]{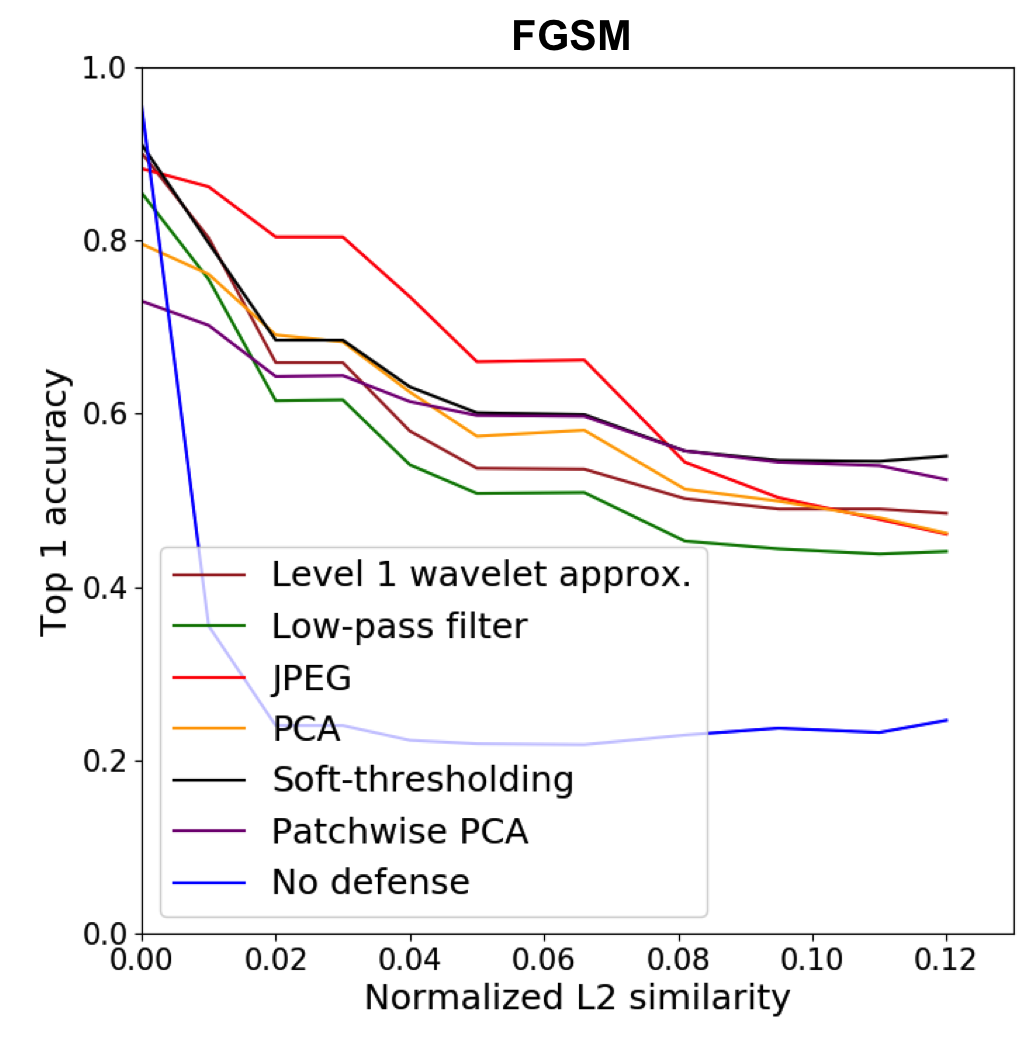}
\includegraphics[width=6.3cm]{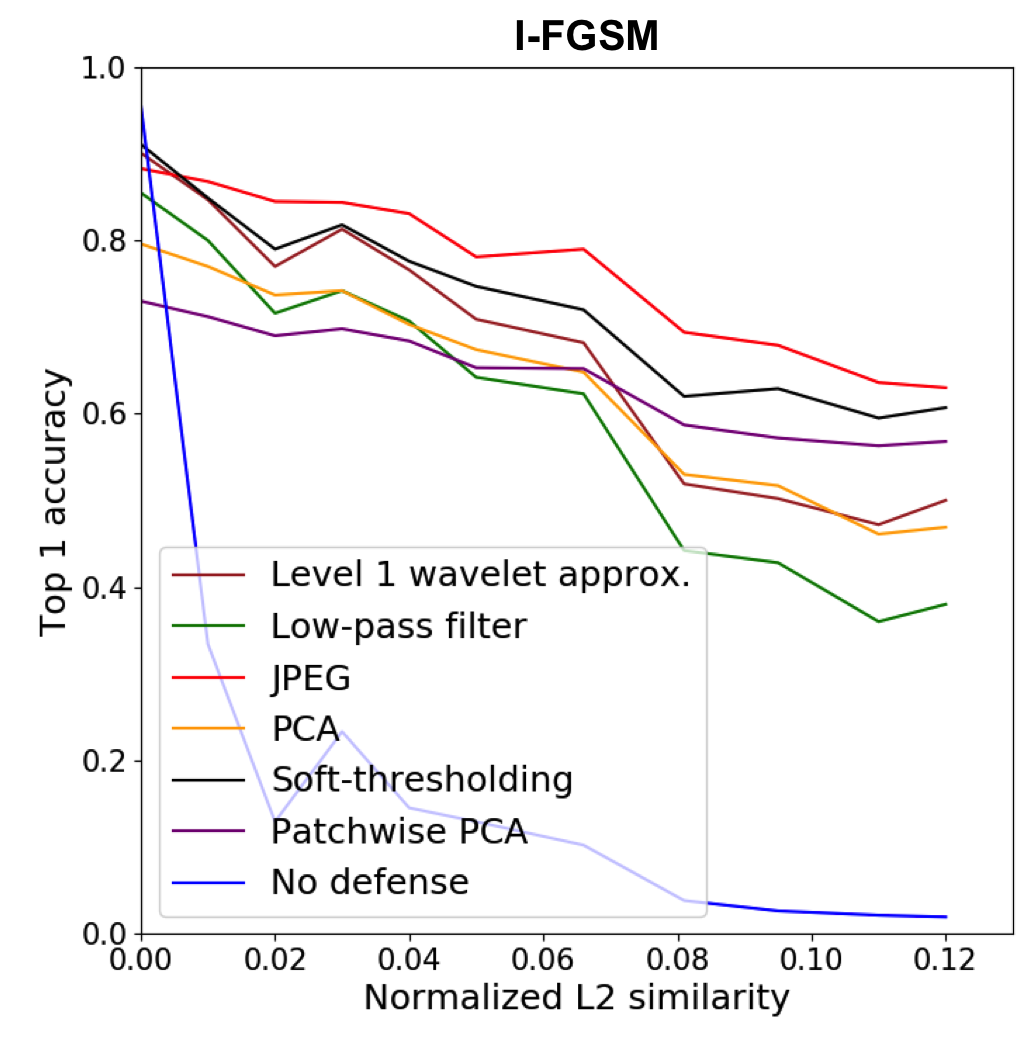}
\includegraphics[width=6.3cm]{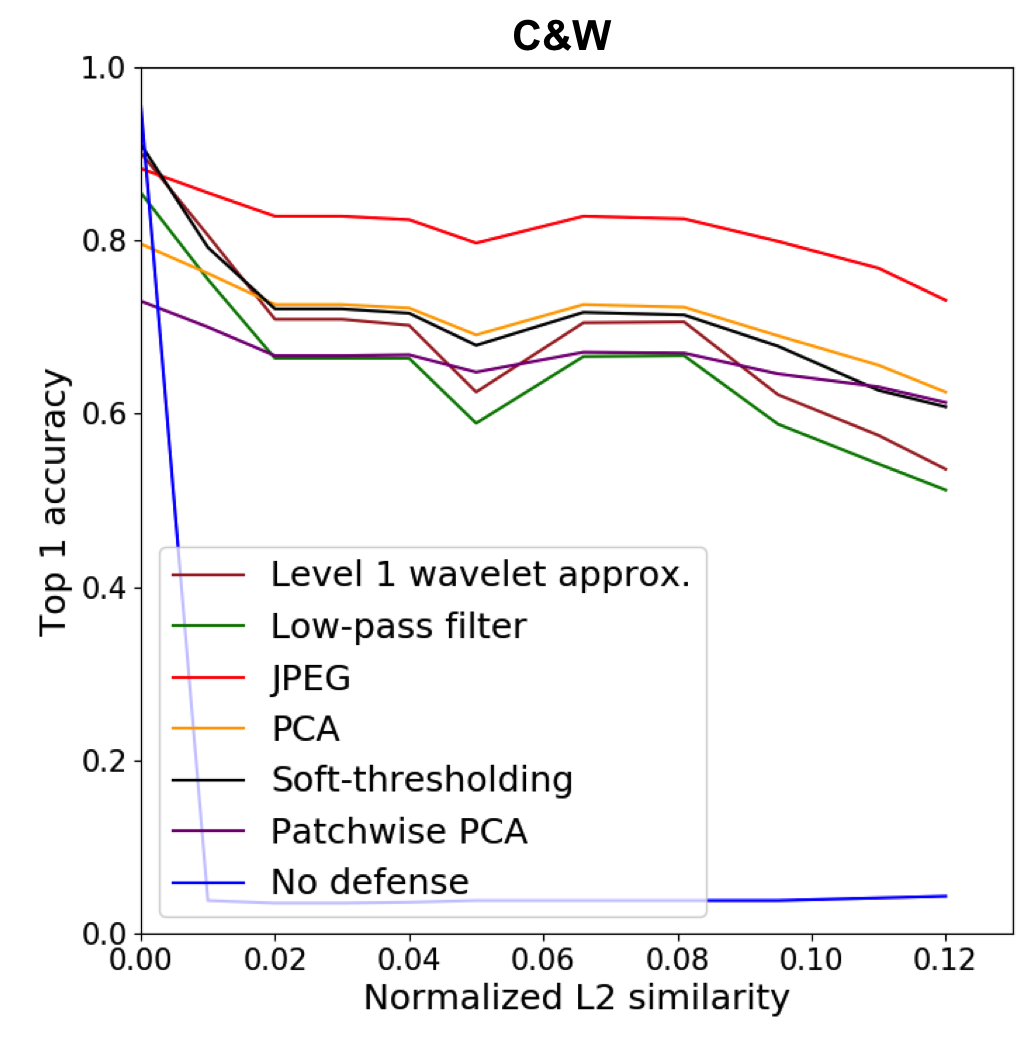}
\caption{Top 1 accuracy of each of the defense methods investigated in this manuscript on FGSM, I-FGSM and C\&W adversarial examples, in a gray-box setting. The examples were generated and tested on Inception-v3.}
\label{fig:defenses}
\end{figure}

In this setting, the attacker has access to the gradients of the target network, but is not aware of the defenses applied.
Specifically, we used the FGSM, I-FGSM and C\&W examples generated for the Inception-v3 network, applied each of the defenses (separately) on each adversarial example, fed them back into Inception-v3, and measured the top 1 accuracy.
The results are shown in Figure~\ref{fig:defenses}.
As can be seen, JPEG denoising performs for the most part as well as or better than all other methods, consistently across all attacks and all perturbation magnitudes. In the FGSM case, as the magnitude of the perturbation gets large, soft-thresholding and patchwise PCA outperform JPEG denoising. 

To complete the evaluation of the defenses in this setting, we also measure the performance of each of the defenses on benign examples, which is shown in Table~\ref{tab:defenses_gray}.
\begin{table}
\centering
\scalebox{0.8}{
\begin{tabular}{ |l|l|}
  \hline
  Defense                                  & Top 1 accuracy    \\\hline\hline
  No defense                               &  0.956            \\\hline
  Low-pass filter                          &  0.855            \\\hline
  PCA                                      &  0.796            \\\hline
  JPEG                                     &  0.883            \\\hline
  Level 1 wavelet approximation            &  0.901            \\\hline
  Soft-thresholding                        &  0.911            \\\hline
  Patchwise PCA                            &  0.73             \\\hline
\end{tabular}
}
\caption{Performance of the defenses on benign examples on Inception-v3. The performance of each defense on adversarial examples is shown in Figure~\ref{fig:defenses}.}
\label{tab:defenses_gray}
\end{table}
The results are consistent with these of Table~\ref{tab:defenses_black}, however the network seems to be more sensitive to the defenses; for example, soft-thresholding yields a 4.5\% decrease in accuracy, comparing to 2.6\% in Table~\ref{tab:defenses_black} and JPEG denoising yields 7.3\% decrease, comparing to 5.5\% in Table~\ref{tab:defenses_black}.


\subsection{White-Box Setting}
\label{sec:whiteBox}
In this setting, the attacker has access to the gradients of the target network, and also has full knowledge of the defense being applied.
Below we present two specific schemes where we utilize this knowledge.
\newline
\textbf{Filtered Gradient Attack (FGA): }
Let $x\in \mathbb{R}^d$, and let $D\in\mathbb{R}^{d\times k}$ be an orthonormal set of basis functions on $\mathbb{R}^d$ (e.g., principal components, complex exponentials or wavelet functions). 
Write $D=[D_\text{retained}, D_\text{filtered}]$, where $D_\text{retained}\in\mathbb{R}^{d\times k_r}$ and $D_\text{filtered}\in\mathbb{R}^{d\times k_f}$ are the subsets of retained and filtered basis functions, respectively, and $k=k_r+k_f$.
We can write $x$ using the basis functions as 
\begin{equation}
x = D_\text{retained}z_r + D_\text{filtered}z_f,\notag
\end{equation}
where $z_r,z_f$ are vectors describing the coefficients of each basis function in the representation of $x$.
Let $J_{\theta,y}(x)$ be the loss of a neural net with parameter $\theta$ for the example $(x,y)$, and let $\nabla_x J(x,y;\theta)\in \mathbb{R}^d$ be its corresponding gradient w.r.t $x$.
A Filtered Gradient Attack  would only modify $z_r$. This can be done by computing the gradient of the loss w.r.t $z_r$, using the chain rule:
\begin{align}
\nabla_{z_r} J_{\theta,y} &=\frac{\partial x}{\partial z_r}\cdot\nabla_x J_{\theta,y} \notag\\
&=D_\text{low}^T\nabla_x J_{\theta,y}\label{eq:grad}.
\end{align}
The gradient $\nabla_{z_r} J_{\theta,y}$ in~\eqref{eq:grad} is defined in the $k_r$-dimensional space of functions in $D_\text{retained}$. 
To map it back to $\mathbb{R}^d$, one should multiply it from the left by $D_\text{retained}$, which defines the adversarial perturbation in the input space $\mathbb{R}^d$ as 
\begin{equation}
D_\text{retained}D_\text{retained}^T\nabla_x J_{\theta,y}(x)\label{eq:lowFreqGrad}.
\end{equation}

Equation~\eqref{eq:lowFreqGrad} simply describes a filtered gradient $\nabla_x J_{\theta,y}(x)$, hence the attack name. 
Some cases of special interest are where the retained basis functions correspond to low frequency terms, first principal components, or coarse wavelet functions; in these cases the FGA perturbation is smoother than usual adversarial perturbations.
More generally, we can apply any of the de-noising procedures in this manuscript on the gradient, to obtain a smooth adversarial perturbation.
In this section we apply each of the de-noising procedures in this manner within a FGSM attack, which results in the following procedure, applied to a raw image $x$:
\begin{enumerate}
\item Forward-propagate $x$ through the net, and compute its loss.
\item Obtain the gradient $\nabla_x J_{\theta,y}(x)$ of the loss w.r.t $x$.
\item De-noise the gradient to get $den(J_{\theta,y}(x))$.
\item $x_\text{adversarial} = \text{clip}(x + \epsilon\cdot \sgn(den(J_{\theta,y}(x)))$.
\end{enumerate} 
\textbf{Backward Pass Differentiable Approximation (BPDA): }
This attack was proposed in~\cite{athalye2018obfuscated} for cases where it is hard or impossible to compute the gradient of a pre-processor which is applied as defense. 
Specifically, we can view the de-noising defense as the first layer of the neural net, which performs pre-processing of the input. When this pre-processing is differentiable, standard attacks can be utilized. 
When it is impossible to compute the gradient of the pre-processing, Athalye et al. propose to approximate it using the identity function, which they justify since the pre-processing step computes a function $g(x)\approx x$.
We apply this logic within a FGSM framework, which results in the following procedure, applied to a raw image $x$:
\begin{enumerate}
\item De-noise $x$ using any of the defense methods to get $den(x)$
\item Forward-propagate the $den(x)$ through the net, and compute its loss.
\item Obtain the gradient $\nabla_x J_{\theta,y}(den(x))$ of the loss w.r.t to the de-noised image.
\item $x_\text{adversarial} = \text{clip}(x +\epsilon\cdot\sgn(\nabla_x J_{\theta,y}(den(x)))$.
\end{enumerate}

\begin{figure}[t!]
  \centering
\includegraphics[width=6.8cm]{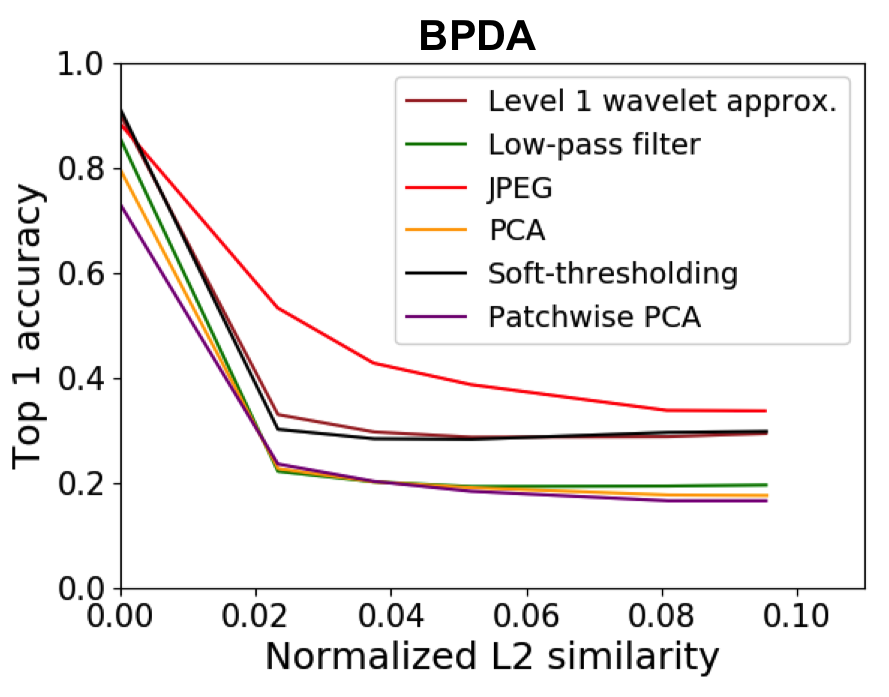}
\includegraphics[width=6.8cm]{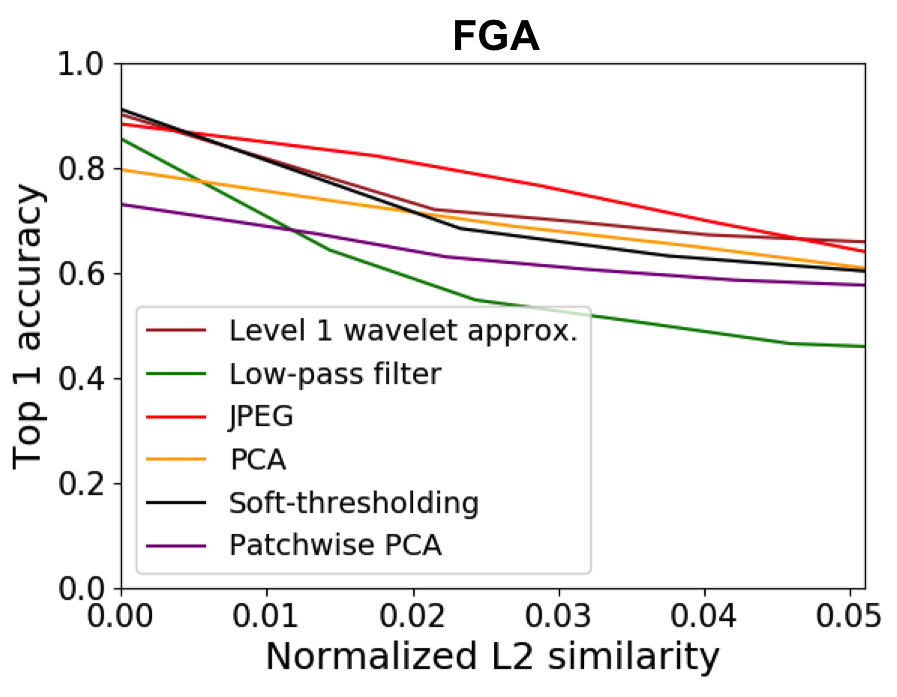}
\caption{Top 1 accuracy of each of the defense methods investigated in this manuscript in a white-box setting, using BPDA and FGA attacks.}
\label{fig:defenses_white}
\end{figure}

We tested the FGA and BPDA using all defense techniques considered in this work; the results are shown in Figure~\ref{fig:defenses_white}.
As can be seen, JPEG appears to be the most successful defense among all tested defenses, under both attack schemes.

%
%


\section{Conclusions and Future Work}
\label{sec:conclusions}
We explored various pre-processing techniques as defenses against adversarial attacks by applying them as test-time pre-processing procedures and measuring their performance under gray, black and white-box settings. 
Our results empirically show that in a black-box setting, JPEG compression and soft-thresholding perform best, while the former outperforms all other tested defenses in gray-box and the two white-box setting considered.
In addition, we proposed the Filtered Gradient Attack, a novel white-box attack scheme, where only components corresponding to a pre-defined basis functions are changed. A special case of FGA is a 'low-frequency' attack.



\bibliography{references}
\bibliographystyle{iclr2018_conference}
\end{document}